\documentclass[conference]{IEEEtran}
\usepackage{amsmath}
\usepackage{amsfonts}
\usepackage[dvipsnames]{xcolor}
\usepackage{listings}
\usepackage{graphicx}
\usepackage{booktabs}
\usepackage{url}
\definecolor{backcolour}{rgb}{0.95,0.95,0.92}
\definecolor{codegreen}{rgb}{0,0.6,0}

\lstdefinestyle{myStyle}{
    backgroundcolor=\color{backcolour},   
    commentstyle=\color{codegreen},
    basicstyle=\ttfamily\footnotesize,
    breakatwhitespace=false,         
    breaklines=true,                 
    keepspaces=true,                 
    numbers=left,       
    numbersep=5pt,                  
    showspaces=false,                
    showstringspaces=false,
    showtabs=false,                  
    tabsize=2,
}

\title{Information Flow in Graph Neural Networks: \\ A Clinical Triage Use Case}
\author{Author 1, Author 2, and Author 3}
\author{V\'ictor Valls, Mykhaylo Zayats, Alessandra Pascale \\
IBM Research Europe -- Dublin}

\begin{document}
\maketitle

\begin{abstract}
Graph Neural Networks (GNNs) have gained popularity in healthcare and other domains due to their ability to process multi-modal and multi-relational graphs. However, efficient training of GNNs remains challenging, with several open research questions. In this paper, we investigate how the flow of embedding information within GNNs affects the prediction of links in Knowledge Graphs (KGs). Specifically, we propose a mathematical model that decouples the GNN connectivity from the connectivity of the graph data and evaluate the performance of GNNs in a clinical triage use case. Our results demonstrate that incorporating domain knowledge into the GNN connectivity leads to better performance than using the same connectivity as the KG or allowing unconstrained embedding propagation. Moreover, we show that negative edges play a crucial role in achieving good predictions, and that using too many GNN layers can degrade performance.

\end{abstract}

%%%%%%%%%%%%

\section{Introduction}

Machine learning algorithms were originally designed to work with data that can be represented as a sequence (e.g., text) or grid (e.g., images). However, these data structures are inadequate for modeling the data of modern applications. For instance, in digital healthcare, a patient's electronic health record (EHR) can include numerous elements, such as demographic information, medical and medication history, laboratory results, etc. One way to model data with arbitrary structure is to use a Knowledge Graph (KG): a graph where nodes represent pieces of information and the edges indicate how the information pieces relate to one another. 

\begin{figure}
\centering
\includegraphics[width=0.85\columnwidth]{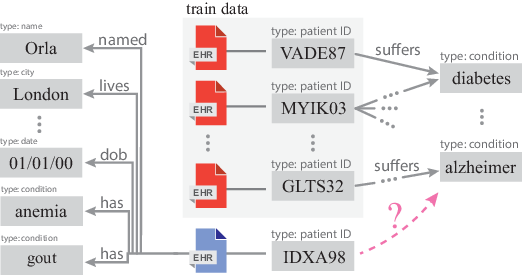}
\caption{Example of a Knowledge Graph (KG) representing the medical record of a patient (Orla). The gray boxes represent the nodes, and the arrows the edges. The dashed and colored arrow with the question mark is the link we would like to predict.  }
\label{fig:intro_example}
\end{figure}

Many learning problems on KGs can be cast as predicting links between nodes. Fig.\ \ref{fig:intro_example} shows an example of a chronic disease prediction problem on a KG. The patient (IDXA98) is connected to its EHR (with the patient’s information such as name, dob, medical conditions, etc.), and the goal is to predict to which chronic disease nodes the patient is connected (colored arrow in Fig.\ \ref{fig:intro_example}). Making such a prediction is possible by analyzing the EHR of other patients with \emph{known} chronic diseases. Another example of a link prediction problem on a KG is when a patient is already diagnosed with a disease (e.g., SARS-CoV-2), and the goal is to find the most effective drug/treatment to help the patient recover \cite{ZSM+20}.

While there exist several methods for predicting edges on graphs \cite{BUG13,XHH+15, ZCZ+20},  Graph Neural Networks (GNNs) have emerged as one of the most widely used techniques.  In brief, GNNs were developed in parallel by two communities: \emph{geometric deep learning} and  \emph{graph representation learning}. The first community focused on applying neural networks for prediction tasks on graph data, while the latter community concentrated on learning low-dimensional vector representations of the nodes and edges in the graph~\cite{Zhou2020}. Current GNNs approaches combine the efforts of both communities and include important extensions such as the ability to handle multi-modal and multi-relational data~\cite{SKB+18,Ahmedt21}. 

GNNs' ability to process multi-modal and multi-relational graphs boosted their popularity in various domains, including healthcare. 
%Another active area of GNN adoption for digital health is related to EHR based predictions. \bc{I don't understand the last sentence.} 
Some applications include the prediction of hospital readmission~\cite{Siyi2023, Golmaei2021}, chronic diseases~\cite{Baribiero2021},  and  ICU mortality~\cite{Zhu2021, Choi2020}.
However, despite their popularity, efficient training of GNNs remains challenging. 
Previous work has primarily focused on designing new architectures for embedding aggregation \cite{SKB+18}, with little emphasis on how embedding information should be exchanged in the network. For instance, the works in \cite{SKB+18, HYL17} suggest that the GNNs' connectivity---which determines how nodes receive information from their neighbors---should align with the connectivity of the KG. However, there are cases where it could be advantageous to explore more complex GNN connectivities that are tailored to the specific task at hand. For example, exchanging embeddings based on the KG connectivity depicted in Fig.\ \ref{fig:intro_example} precludes medical conditions from influencing patient embeddings. Yet, incorporating such interactions can be beneficial in tasks such as chronic disease prediction, where it is essential to capture the patients' existing medical conditions (e.g., hyperglycemia for predicting diabetes) in their embeddings.%However, incorporating such interactions can be advantageous in tasks like chronic disease prediction, where it is important to capture the patients' existing medical conditions (e.g., hyperglycemia for predicting diabetes) in their embeddings.

In this paper, we investigate how the flow of embeddings within a GNN affects the prediction of links in a clinical triage use case. The paper makes the following contributions:

\begin{itemize}
\item [1)] 
We present a mathematical model for predicting links on KGs with GNNs, where we cast the prediction task as an optimization problem and leverage GNNs as an algorithmic tool to solve it (Sec. \ref{sec:mathematical_model}). This model emphasizes that the GNN design parameters, such as the GNN connectivity, can be decoupled from the underlying structure of the graph data (i.e., the KG).

\item [2)] We show how to map the link prediction optimization to a program in PyG (Sec.\ \ref{sec:link_prediction_pyg}) and study how the GNN parameters affect the link prediction accuracy in a clinical triage use case (Sec. \ref{sec:usecase}). Our findings suggest that a GNN connectivity that considers domain knowledge is more effective than just using the connectivity of the graph data, and that allowing embeddings to flow in any direction may result in poor performance (Sec.\ \ref{sec:gnn_connectivity}). Additionally, we demonstrate that negative edges play a crucial role in achieving good predictions (Sec.\ \ref{sec:neg_edge_exp}), and that using too many GNN layers can degrade performance (Sec.\ \ref{sec:emb_layer_exp}).%Our results show that using a GNN connectivity that considers the data interactions is more effective than the connectivity of the graph data. Additionally, our study highlights the importance of incorporating domain knowledge through negative links and shows that using many GNN layers can compromise performance.
\end{itemize}

%%%%%%%%%%%%%%%%%
\section{Link Prediction Model}
\label{sec:mathematical_model}
%%%%%%
\subsection{Multi-relational Knowledge Graph (KG)}
\label{sec:mrel_kg}
A multi-relational Knowledge Graph (KG) is a graph with $n$ nodes and $m$ \emph{directed} links, where each link is associated with a relation $r$ that represents the type of connection between the nodes. For instance, in the semantic triple \emph{patient} (node) \emph{suffers from} (relation) \emph{anemia} (node), the relation \emph{suffers from} indicates the type of connection between the node \emph{patient} and the node \emph{anemia}. The nodes in the graph are also associated with a type or class, e.g., the node \emph{anemia} can be of the type \emph{medical condition}.

Besides the links' relation, a link can be \emph{positive}, \emph{negative}, or \emph{unknown}. A positive link indicates the two nodes are connected, while a negative link implies no connection. For example, if a patient has tested positive for diabetes, there will be a  (positive) link between the patient's node and the diabetes node in the KG. Conversely, if the patient has tested negative, a (negative) link will indicate that such a connection does not exist.  Unknown links, as the name suggests, are links whose existence is unknown from the data. This is the type of link that we would like to predict.

%%%%%%%
\subsection{Link prediction as an optimization problem}

We can model a link prediction problem in a KG as follows. Every node $i \in \{1,\dots, n\}$ is associated with a feature vector $e_i \in \mathbf R^d$, which is also known as the \emph{node's embedding}, or \textit{embedding} for short. Similarly, every link is associated with a relation matrix $W_r \in \mathbf R^{d \times d}$, $r \in \{1,\dots,R\}$. 
Next, for every pair of nodes $(i,j)$ and relation $r$, we define the links' ``score'' as
\begin{align}
x^{(r)}_{ij} : = f(e_i,  W_r, e_j) 
\label{eq:score}
\end{align}
where  $f$ is a function that takes $e_i$, $W_r$, and $e_j$ as inputs and returns a real number in the interval $[0,1]$.\footnote{For example,  $f(z): =  (1+ \exp(-z))^{-1}$ and $z := e_i^T  W_r e_j$.} Similarly, for every link connecting nodes $(i,j)$ with a relation $r$, we define the labels
\begin{align*}
y^{(r)}_{ij} = \begin{cases}
1 & \text{there is a relation $r$ from node $i$ to $j$}, \\
0 & \text{there is \underline{not} a relation $r$ from node $i$ to $j$} .
\end{cases}
\end{align*} 

With the above model, we can formulate the optimization problem
\begin{align}
\begin{tabular}{ll}
$\underset{e_i , W_r}{\text{minimize}}$ & $ \mathcal L(\mathbf x, \mathbf y) $
\end{tabular}
\label{eq:optimization-problem}
\end{align}
where  $ \mathcal L : \mathbf R^m \times \mathbf R^m \to \mathbf R$, $\mathbf x = (x^{(r)}_{ij}) \in \mathbf R^m$,  $\mathbf y \in \{ 0,1 \}^{m}$. The role of the loss function $ \mathcal L $ is to penalize vector $\mathbf x$ being different from vector $\mathbf y$ component-wise.\footnote{For example, $\mathcal L$ can be $\|\mathbf x - \mathbf y \|_2$, i.e., the $\ell_2$-norm.  }
Namely, by minimizing $ \mathcal L$ in \eqref{eq:optimization-problem}, we are finding the nodes' embeddings $e_i$ and the matrices $W_r$ such that the score $x_{ij}^{(r)}$ is equal to (or, close to) the label $y_{ij}^{(r)}$, which indicates the presence of a positive/negative link.

%%%%%%%%%%
\subsection{Solving the link prediction problem with a GNN}

\begin{figure}
\centering
\includegraphics[width=0.75\columnwidth]{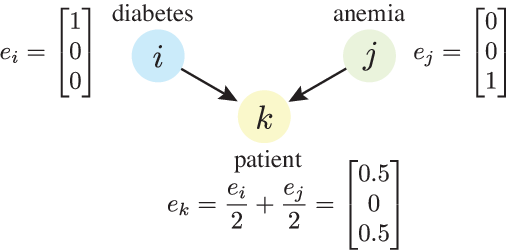}
\caption{Toy example of how the embedding of a patient is the linear combination of two medical conditions embeddings.}
\label{fig:simple_nn}
\end{figure}

GNNs tackle the optimization problem \eqref{eq:optimization-problem} by computing nodes' embeddings based on their connectivity patterns with neighboring nodes. To illustrate the concept, we show in Figure \ref{fig:simple_nn} a toy example where a ``patient'' feature vector is built with the patient's medical conditions embeddings. In particular, the feature vectors of nodes $i$ and $j$ (medical conditions) are combined \emph{linearly} to obtain the embedding of node $k$ (a patient).

\begin{figure}
\centering
\includegraphics[width=0.7\columnwidth]{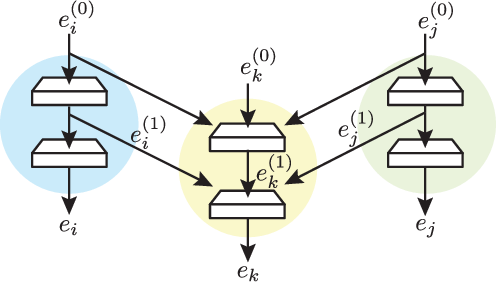}
\caption{Example of a GNN with three nodes and two NN layers per node. Vectors $e_i^{(0)}$, $e_j^{(0)}$, $e_k^{(0)} $ are the initial embeddings of nodes $i$, $j$, $k$, i.e., the input in the first NN layer. Vectors   $e_i^{(2)} = e_i$, $e_j^{(2)} = e_j$, $e_k^{(2)} = e_k$ are the embedding outputs of layer 2. }
\label{fig:nn}
\end{figure}

GNNs combine the neighbors' embeddings by using multiple \emph{non-linear} functions. Fig.\ \ref{fig:nn} shows an example of how a GNN uses multiple layers (i.e., functions) to combine the embeddings. Each layer $l \in \{1,\dots, L\}$ fuses the feature vector in the $(l-1)$-th layer of the node with embeddings of its neighbors, and the output is passed to the next layer where the process is repeated.
 
The GNN connectivity depends on how nodes are connected in the KG (but is not necessarily the same), and it determines how the embedding information propagates. Designing a GNN connectivity that enables efficient learning is use-case dependent as it requires knowing how nodes should interact.  In Sec.\ \ref{sec:gnn_connectivity}, we will show how different GNN connectivities affect the link-prediction performance for a clinical triage use case. 

%Finally, it is worth mentioning that one notable advantage of GNNs is their ability to efficiently compute the weights used for combining embeddings. Moreover, GNN functions can often run on a GPU, significantly accelerating the optimization process.
%

\subsection{Link prediction}

Predicting an (unknown) link/relation in a KG consists of evaluating \eqref{eq:score} with the embeddings and relation weights learned during the training. For example, suppose we have a patient  and want to predict whether the patient may be \emph{infected} by COVID-19. Then, we use $e_{\text{patient}}$, $W_{\text{infected}}$,  and $e_\text{COVID-19}$  in  \eqref{eq:score}, and if the score is larger than a confidence threshold (e.g., $0.9$), we can determine a positive link exists.

Often, we need to predict links that connect nodes not seen during the training. In that case, we need to first compute the embeddings of the new nodes by combining their initial embeddings\footnote{The input in the first NN layer. See Fig.\ \ref{fig:nn}.} with the embeddings of their neighbors seen in the training. For example, a new patient may be connected to nodes that appeared during the training (e.g., fever, headache, cough). Then, the embedding of the unseen node (i.e., the patient) is calculated with the embeddings of the neighboring nodes.

%%%%%%%%%%%%%

\section{Link Prediction in PyG}
\label{sec:link_prediction_pyg}

This section presents how to implement the link prediction optimization in Sec.\ \ref{sec:mathematical_model} with PyG \cite{pyg}---a python library for GNNs built upon PyTorch.  We follow a similar approach as in the PyG tutorial for node classification \cite{PyG-introductory-example}. 

\subsection{Creating the KG and tensors}
The first step is constructing a KG with the format in Table \ref{table:kg}. Each row in the table corresponds to a subject-relation-object ``triple'' indicating how nodes are connected. Recall the edges in the KG are directed, where the \emph{subject} and \emph{object} are the \emph{source} and \emph{target} nodes, respectively.\footnote{The names in the subject and object columns identify a single node in the KG. That is, there cannot be multiple nodes called \emph{London} referring to different cities, e.g., England (UK), Ontario (Canada), Texas (USA), etc.}
 The column \emph{link type} indicates whether such a link is positive (\texttt{True}) or negative (\texttt{False}), and the columns \emph{sub.\ type} and \emph{obj.\ type} indicate---as the names suggest---the types of nodes. Having different node types is useful, for example, to control how the embedding information flows in the GNN. In Sec.\ \ref{sec:gnn_connectivity}, we will show how different GNN connectivities affect the link prediction performance. 
 
The second step is to map the KG to a PyG \texttt{data} object that contains the nodes' initial embeddings (\texttt{data.x}), the network connectivity (\texttt{data.edge\_index}), the types of relations (\texttt{data.edge\_type}), and the labels (\texttt{data.y}) that indicate whether the links are positive or negative.  Fig.\ \ref{fig:tensor_example} shows how to map the KG in Table \ref{table:kg} to a \texttt{data} object, where the nodes and relations have unique IDs (integers).

\begin{table}
\caption{Example of a table that represents the connectivity of a KG. Each row is an edge in the KG. }
\label{table:kg}
\centering
\begin{tabular}{| l | l | l | l | l | l | }
\hline
subject & relation & object & sub.\ type & obj.\ type & link type\\
\hline
Orla	& born & London & person	& city & \texttt{True} \\
Orla	& has & cholesterol & person	& disease & \texttt{True} \\
Paul	& born & New York & person & city & \texttt{True} \\
Paul & has & diabetes & person & disease & \texttt{False} \\
\hline
\end{tabular}
\end{table}
\begin{figure}

\centering
\small
\setlength{\jot}{0pt}
\begin{align*}
\texttt{data.edge\_index = tensor([} & \texttt{[0, 2],} \\
& \! \! \! \texttt{ [0, 3],} \\
& \! \! \! \texttt{ [1, 4],} \\
& \! \! \! \texttt{ [1, 5]])}
\end{align*}
\begin{align*}
\qquad \texttt{data.edge\_type = tensor([0, 1, 0, 1])} \\
\qquad \texttt{data.y = tensor([1, 1, 1, 0])}
\end{align*}
\vspace{-2em}
\caption{Example of a torch tensor that maps the KG in Table \ref{table:kg} with unique IDs. Each row coresponds to a row in Table \ref{table:kg} where nodes' IDs are assigned sequentially: Orla (0), Paul (1), London (2), cholesterol (3), New York (4), diabetes (5). The relations' IDs are also assigned sequentially: born (0), has (1). }
\label{fig:tensor_example}
\end{figure}

%\begin{figure}
%
%\centering
%\small
%\setlength{\jot}{0pt}
%\begin{align*}
%\texttt{tensor([} & \texttt{[0, 0, 2],} \\
%& \! \! \! \texttt{ [0, 1, 3],} \\
%& \! \! \! \texttt{ [1, 0, 4],} \\
%& \! \! \! \texttt{ [1, 1, 5]])}
%\end{align*}
%\vspace{-2em}
%\caption{Example of a torch tensor that maps the KG in Table \ref{table:kg} with unique IDs. Each row coresponds to a row in Table \ref{table:kg} where nodes' IDs are assigned sequentially: Orla (0), Paul (1), London (2), cholesterol (3), New York (4), diabetes (5). The relations' IDs are also assigned sequentially: born (0), has (1). }
%\label{fig:tensor_example}
%\end{figure}

%%
\subsection{GNN model} 

\begin{figure}[t!]
\centering
\includegraphics[width=\columnwidth]{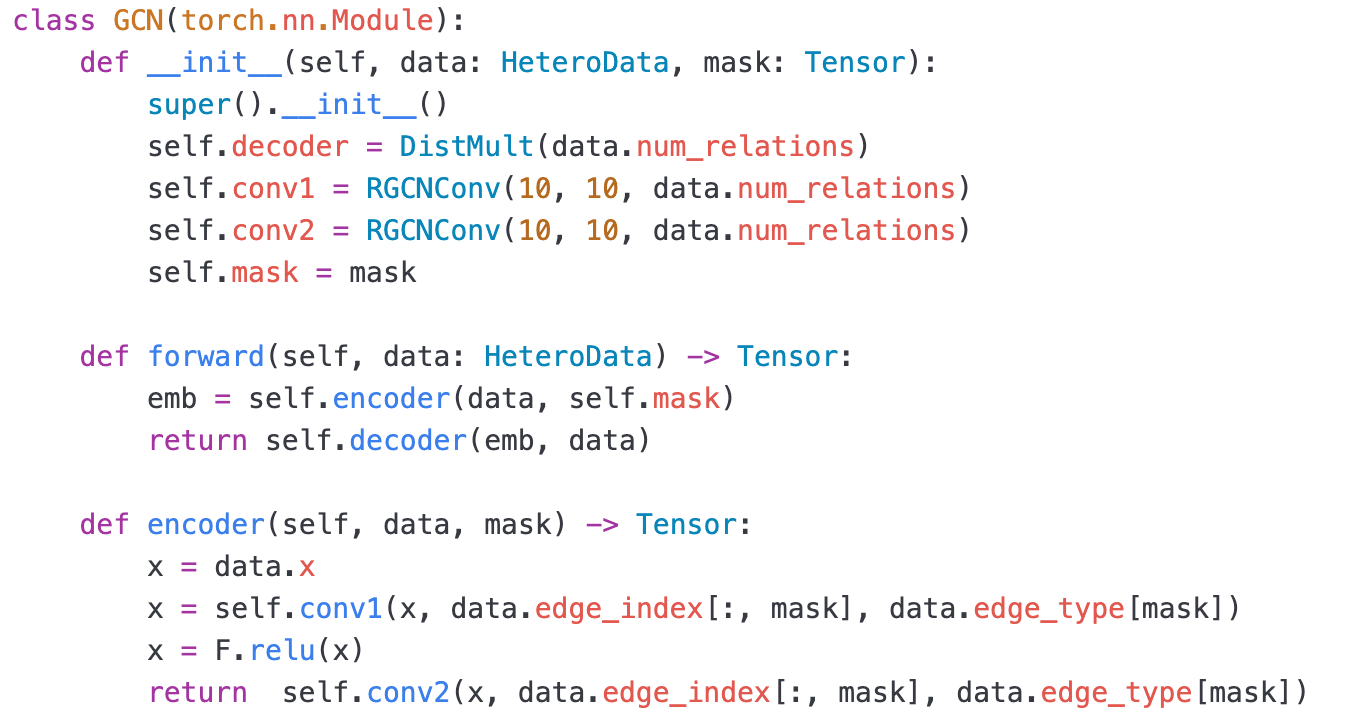}
\caption{Example of a torch module to implement a GNN with two RGCNConv layers. The layer architecture was proposed in \cite{SKB+18} and is available directly in PyG.}
\label{fig:GNN_code}
\end{figure}

The GNN model consists of two core parts: (i) the generation of the embeddings (i.e., encoder) and (ii) the scoring function (i.e., decoder). Fig.\ \ref{fig:GNN_code} shows an example of a GNN model with two RGCNConv layers \cite{SKB+18}.\footnote{RGCNConv is a type of layer/function to compute the embeddings.} The first part is to initialize the scoring function\footnote{The weights $W_r$ are part/defined in the scoring function. } and the NN functions that will generate the embeddings. % the relation weights (i.e., $W_r$), and the initial embeddings (i.e., $e_i^{(0)}$, $i \in \{1,\dots,n\}$).\footnote{See the example in Fig.\ \ref{fig:nn}.} 
The initial embeddings ($\texttt{data.x}$) in the encoder and can be set manually---e.g., using a pre-trained natural language model that maps the node's name (i.e., a string) to a vector of a fixed size (e.g., as in \cite{LYK+20})---or they can be variables in the optimization. 

The forward function computes the nodes' embeddings and scores for every link in the KG. The embedding information is obtained with a communication mask\footnote{A mask is a  vector of booleans that selects which edges (i.e., the tensor's rows) to use.} that controls how the nodes propagate their embeddings to their neighbors in the GNN, i.e., the GNN connectivity.

\subsection{Training the model}

\begin{figure}[t!]
\includegraphics[width=\columnwidth]{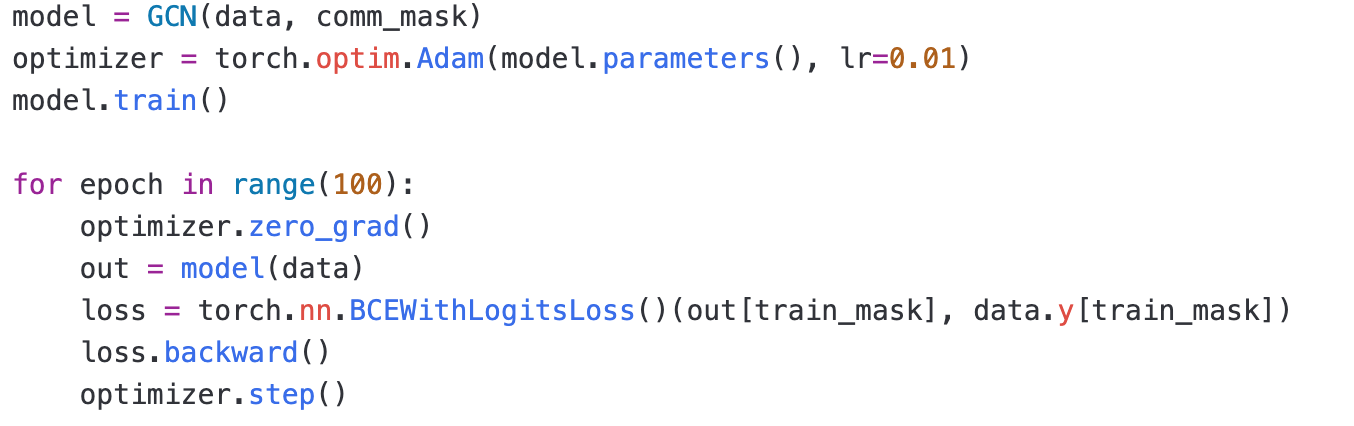}
\caption{Example of the optimization procedure using the steps in \cite{PyG-introductory-example}. }
\label{fig:code_opt}
\end{figure}

The training of the GNN is shown in Fig.\ \ref{fig:code_opt}. The process follows the steps in the tutorial for node classification in \cite{PyG-introductory-example} with two differences. The first one is that we can control the embedding communication in the GNN, which affects how the initial embeddings and the embedding in intermediate layers are combined. The second difference is that we can filter the edges for which we want to evaluate the loss function, which is useful to specialize the model to the types of links we would like to predict.\footnote{This is similar to the mask used in \cite{PyG-introductory-example} to filter the training data.  }

\subsection{Link prediction}

The link prediction task consists of calling the function \texttt{model(data)} where \texttt{data} includes the links we would like to predict. For instance, if we want to predict if the node \emph{Orla} is connected to the node \emph{diabetes} with the relation type \emph{has} (with the mapping used in the example in Fig.\ \ref{fig:tensor_example}), we need to add the \texttt{tensor([0, 5])} to \texttt{data.edge\_index} and \texttt{tensor([1])} to \texttt{data.edge\_type}. 

To predict links connecting nodes not seen in the training, we must first assign the new nodes unique IDs and generate their embeddings using the same function used in the training. However, the GNN communication mask employed to create the embeddings must prevent the unseen/new nodes from affecting the embeddings of the nodes seen in the training. %For example, the mask could remove all the links from the new/unseen nodes to the nodes seen in the training. 

%%%%%%%%%%
\section{Use case: Clinical triage with Synthea}
\label{sec:usecase}

This section presents a numerical evaluation of the GNN for clinical triage with the Synthea dataset generator \cite{WKN+18}. The experiments' goal is to illustrate how (i) the GNN parameters and (ii) the domain knowledge affect the link prediction accuracy. In particular, we study different GNN connectivities (Sec.\ \ref{sec:gnn_connectivity}), embedding sizes, and number of GNN layers (Sec.\ \ref{sec:emb_layer_exp}), and the importance of negative edges in the construction of the KG (Sec.\ \ref{sec:neg_edge_exp}).

\subsection{Use case and dataset overview}
\subsubsection{Use case} The \emph{clinical triage} problem involves determining the appropriate course of care when a patient presents with symptoms or medical conditions at the first point of contact. This includes deciding whether the patient requires immediate attention from a healthcare professional, such as in an emergency situation (e.g., a heart attack). 

\subsubsection{Dataset} Synthea is a \emph{synt}hetic \emph{hea}lthcare dataset generator that simulates realistic patient medical records. The generated patient records include a variety of information such as demographics, medical history, medications, allergies, and encounters with healthcare providers. The resulting data is designed to be representative of the United States population in terms of age, gender, and ethnicity, and it includes data on over 10 million synthetic patients. 

For the clinical triage problem, we access the patient's medical records generated by Synthea and extract the patient's medical conditions and encounters. Each encounter is associated with conditions (e.g., diabetes) and observations (e.g., fever), and belongs to a class that corresponds to one of the following care actions:  \emph{wellness}, \emph{inpatient}, \emph{outpatient},  \emph{ambulatory}, and \emph{emergency}. The goal of  clinical triage problem is: Given a patient's medical encounter with some medical conditions and observations, determine the type of care action the patient should receive, i.e., to which care action node should the encounter be connected.

\begin{figure}[t!]
\centering
\begin{tabular}{cc}
\includegraphics[width=0.45\columnwidth]{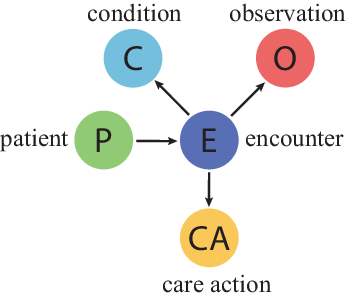} & 
\includegraphics[width=0.45\columnwidth]{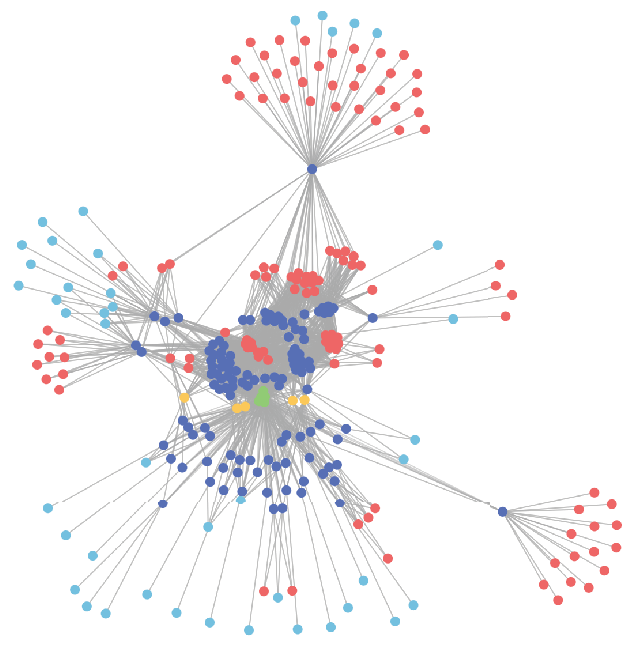} \\
(a) & (b) 
\end{tabular}
\caption{(a)  KG schema generated with Synthea for the clinical triage problem. (b) Example of a graph with 5 patients. The edges in (b) are shown as undirected due to the figure size. }
\label{fig:synthea_kg}
\end{figure}

\subsection{Experiment setup}
\label{sec:experiement_setup}

\subsubsection{KG} 
\label{sec:kg_construction}
We generate a KG for the clinical triage problem with Synthea as shown in Fig.\ \ref{fig:synthea_kg}. There are five types of nodes (\emph{encounter}, \emph{observation}, \emph{condition}, \emph{patient}, and \emph{care action}) connected by four different types of relations (\emph{encounter-careaction}, \emph{encounter-observation}, \emph{encounter-condition}, and \emph{patient-encounter}).  
As a remark, Synthea does not provide information about negative edges. Still, since an encounter can only be connected to one care action, we add negative links\footnote{The negative links are of type \emph{encounter-careaction}.} between an encounter and the other care actions. For example, suppose an encounter has a positive link with the care action \emph{inpatient}. In that case, we add negative links in the KG between the encounter and the care actions \emph{wellness}, \emph{outpatient}, \emph{ambulatory}, and \emph{emergency}.

\subsubsection{GNN} We will introduce the GNN connectivities in the experiment in Sec.\ \ref{sec:gnn_connectivity}. The initial embeddings of the node types \emph{observation}, \emph{condition}, and \emph{care action} are  variables in the optimization problem, while the initial embeddings of the node types \emph{encounter} and \emph{patient} are set to zero. This choice of initial embeddings is because, in the testing, we do not want to infer the embeddings of observations, conditions, or care actions not seen in the training. However, we allow new patients and new encounters. 

\subsubsection{Training parameters} We conducted the numerical experiments with 50/10 patients in the training/testing sets---sampled uniformly at random. All the experiments are run for 50 realizations, where each realization involves a random sample of 50/10 patients from the Synthea generated data.\footnote{A training sample has, on average, 35k edges and 1.9k nodes (50 patients, 1603 encounters, 153 observations, 107 conditions, and 5 care actions).}
The scoring function is DistMult \cite{SKB+18} with weights initialized at uniformly at random, the loss function is logistic regression, and the GNN architecture uses RGCNConv layers \cite{SKB+18}. 
The training is carried out with Adam \cite{KB14} for 1000 epochs with variable learning rate\footnote{The learning rate is equal to 0.1 for the first 100 epochs, 0.01 for the following 600 epochs, and 0.001 for the last 300 epochs. } and weight decay 0.0005.  

\subsection{Experiments}
\label{sec:experiments}

\begin{figure}
\centering
\includegraphics[width=\columnwidth]{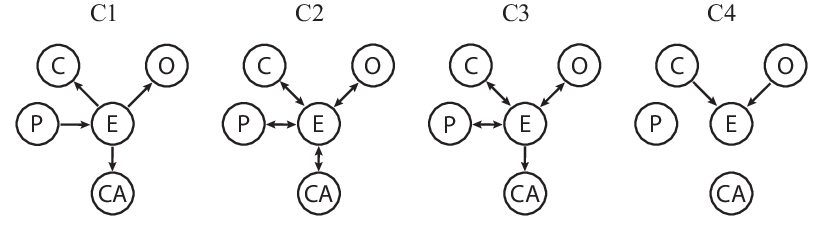}
\caption{The four GNN connectivities used in the experiments in Sec.\ \ref{sec:gnn_connectivity}. Acronyms: condition (C), observation (O), encounter (E), patient (P), and care action (CA). }
\label{fig:four_gnn_connectivities}
\end{figure}

\subsubsection{GNN connectivity}
\label{sec:gnn_connectivity}
This experiment studies how the GNN connectivity for the propagation of the embedding information affects the link prediction performance. We consider the four GNN connectivities shown in Figure \ref{fig:four_gnn_connectivities}. In short, the {C1} connectivity corresponds to the connectivity of the positive links in the KG. The {C2} connectivity is obtained by adding ``reverse'' links in C1.\footnote{The reverse links have relation type \emph{\{sub.\ type\}-\{obj.\ type\}.}} Thus, the embedding information flows in any direction. The {C3} connectivity is as  C2 but without the edge from the node {care action} to {encounter}. Finally, the {C4} connectivity allows only embedding information to flow from observation and condition nodes to encounter nodes. The rationale behind C4 is that the node types observation and condition can be regarded as the ``attributes'' or ``properties'' of an encounter, and therefore the embedding of an encounter should not affect the embeddings of the observation and condition nodes.\footnote{i.e., there should be no edge from an encounter node to an observation/condition node. }%The same rationale applies to the connections between the encounter and care action nodes. 

\begin{table}[t!]
\caption{Percentage of correct care action predictions for different GNN connectivities (experiment in Sec.\ \ref{sec:gnn_connectivity}).}
\label{table:gnn_conn}
\centering
\begin{tabular}{l|llllll}
\bottomrule
 & well. & eme. & amb. & inpat. & outpat. & average \\
 \hline
C1 & 0.73 & 0.00 & 0.27 & 0.00 & 0.00 & 0.43\\
C2 & 0.14 & 0.16 & 0.25 & 0.28 & 0.19 & 0.19 \\
C3 & 0.90 & 0.77 & 0.92 & 0.60 & 0.43 & 0.87\\
C4 &0.96  & 0.76 & 0.94 &0.59  & 0.45 & 0.88 \\
\toprule
\end{tabular}
\end{table}

\begin{figure}[t!]
\centering
\includegraphics[width=0.9\columnwidth]{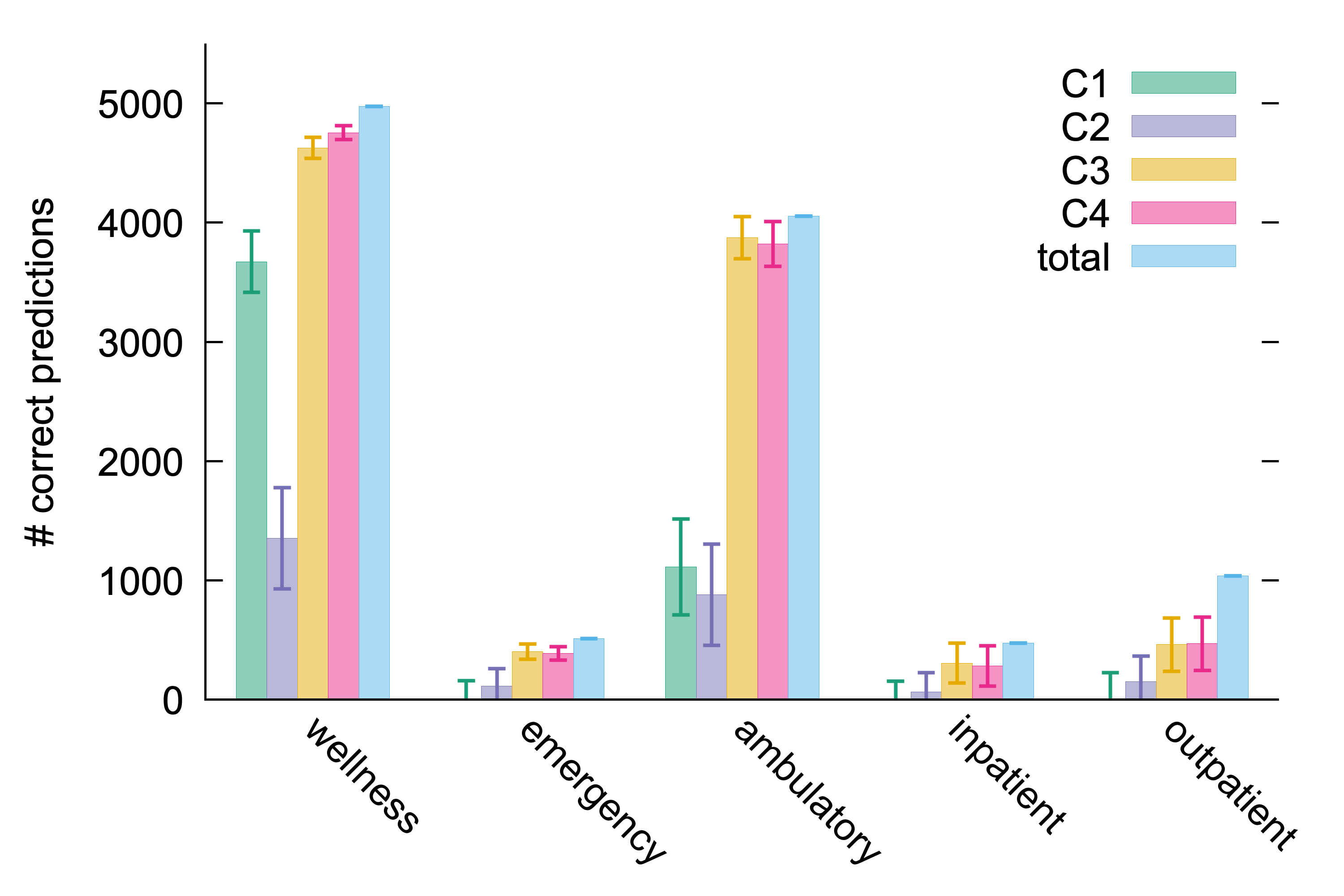}
\caption{Illustrating how the GNNs connectivities in Fig.\ \ref{fig:four_gnn_connectivities} affect the prediction of care actions.  The results are the average of 50 random samples from the Synthea generated dataset. Each sample consists of 50/10 patients in the training/testing sets. The nodes' embeddings have size 5 and the GNN has 2 layers. The total number of predictions per care action is indicated in blue. }
\label{fig:gnn_connecitivty}
\end{figure}

Fig.\ \ref{fig:gnn_connecitivty} shows the number of correct care action predictions for the four GNN connectivities, where \emph{total} (blue bar) indicates the number of care actions of that type in the testing data (ground truth). Observe from the figure that the frequency of care actions is skewed, with \emph{wellness} being the most common and \emph{emergency} being the least common. Regarding correct predictions, the C4 connectivity has the best overall performance, closely followed by the C3 despite C3 having more than twice as many edges as C4 (see also Table \ref{table:gnn_conn}). The C1 and C2 connectivities did not perform well, but for two different reasons. First, the C1 connectivity does not allow the encounter nodes to receive embeddings from the observation and condition nodes, which are the ``characteristics that define an encounter.'' The C2 connectivity fails because we allow the encounter nodes to access information that is not available in the testing. Specifically, the links between care action and encounter nodes do not exist when creating the embeddings in the testing---since they are the links we would like to predict.\footnote{We just want to predict the link from care action to encounter, but the reverse edge from care action to encounter will not exist either in the testing. }

\textbf{Conclusions:} GNN connectivities that may appear intuitive (C1 and C2) do not perform well because (i) the associated KG connectivity does not capture how the nodes' embeddings should interact, and (ii) the training uses links for computing the nodes' embeddings that are not present in the testing. Connecting nodes in every direction (C3) obtains a good performance, but it is slightly outperformed by a bespoke GNN connectivity (C4) that considers only essential connections.

%%%%%%%
\subsubsection{Embedding size and number of GNN layers}
\label{sec:emb_layer_exp}

This experiment investigates the impact of two basic GNN design parameters: The number of layers and the size of the nodes' embedding. The GNN connectivity used here corresponds to the C4 connectivity described in Sec.\ \ref{sec:gnn_connectivity}.

\begin{figure}[t!]
\centering
\begin{tabular}{cc}
\includegraphics[width=0.48\columnwidth]{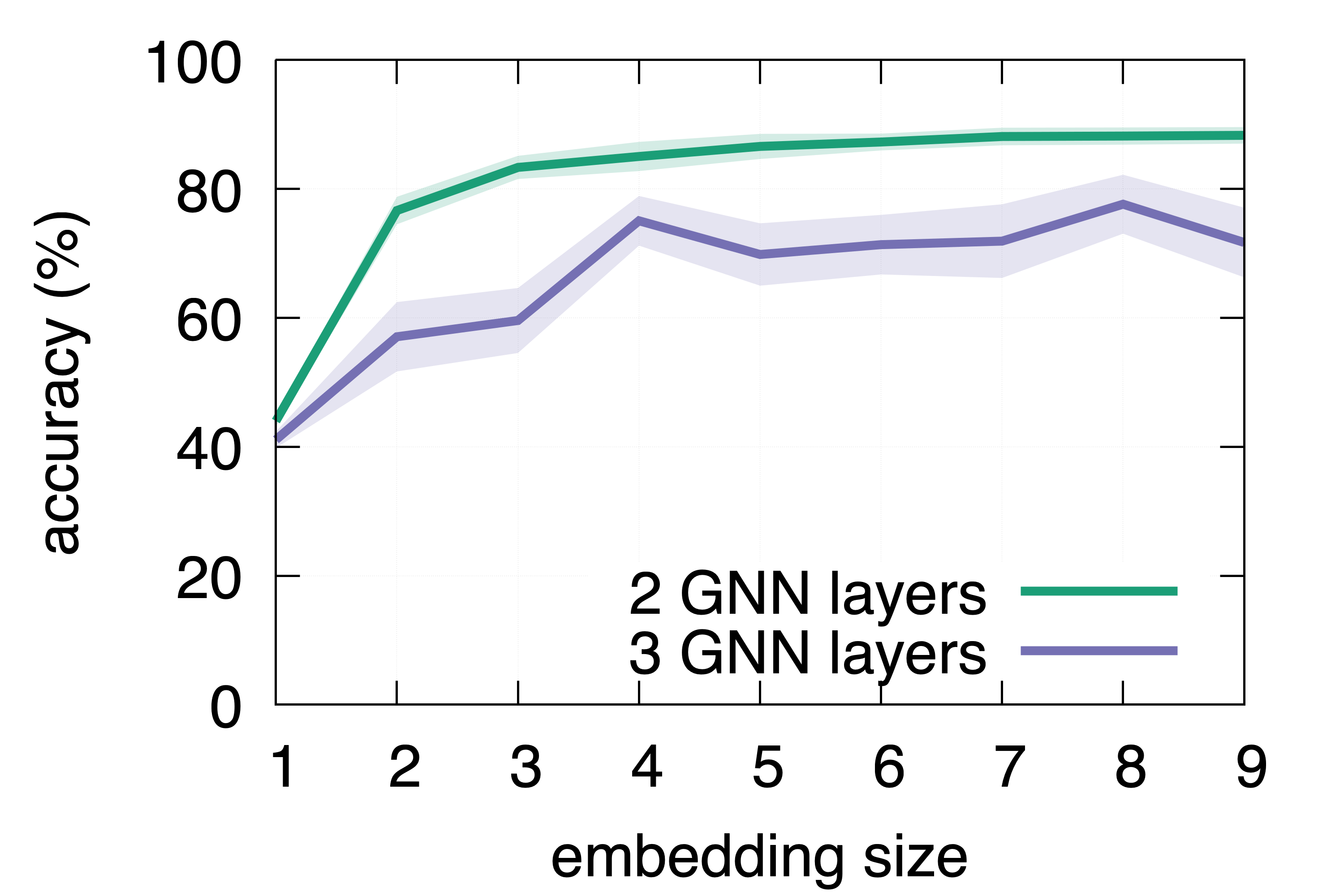} & 
 \includegraphics[width=0.48\columnwidth]{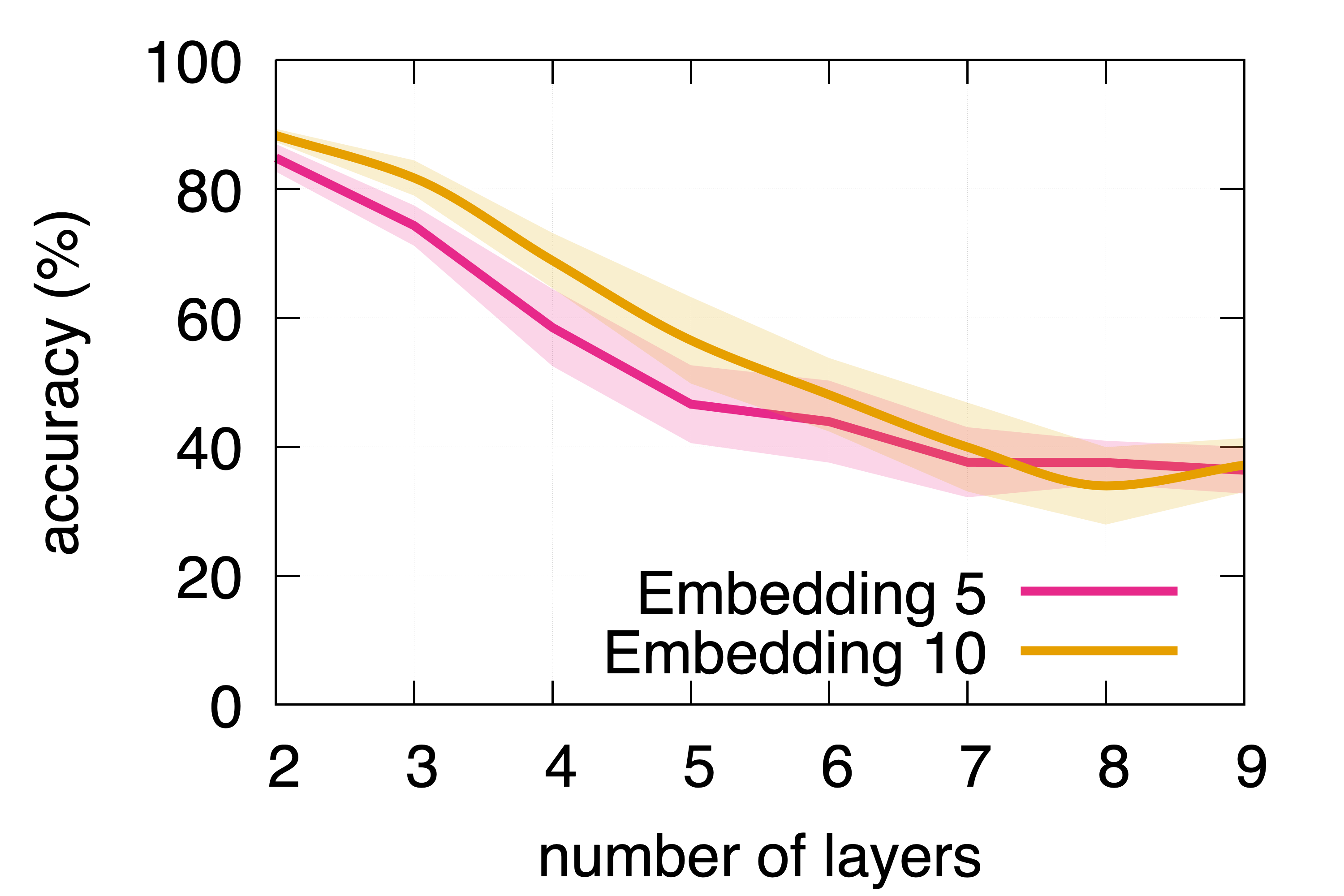} \\
(a) &  (b) 
\end{tabular}
\caption{Illustrating how (a) the nodes' embedding size and (b) the number of GNN layers affects the predictions of care actions. The KG connectivity is C4. The results are the average of 50 random samples of the Synthea dataset. Each sample consists of 50/10 patients in the training/testing. }
\label{fig:embedding_size}
\end{figure}

Fig.\ \ref{fig:embedding_size}a shows the average prediction accuracy of care action as a function of the embedding size for GNNs with 2 and 3 layers. Observe from the figure that, in both cases, the accuracy improves rapidly for embedding sizes ranging from 1 to 3, but beyond that point, the increase in accuracy becomes more gradual. 

Fig.\ \ref{fig:embedding_size}b shows the prediction accuracy as a function of the number of GNN layers when the embedding sizes are fixed to 5 and 10. Observe from the figure that adding more layers decreases the GNN performance, which is in stark contrast to \emph{deep} CNNs, which use many layers. We reckon this behavior is because of the ``over-smoothing'' phenomenon also noted in the literature \cite{CLL+20, CW20}, where adding more layers makes the GNN ``too well connected,'' and therefore, the nodes embeddings become ``too similar.''

 \textbf{Conclusions:} The link prediction performance improves with the embeddings' size, but the improvement gains diminish once the embeddings are large enough. Using a large number of GNN layers has a negative impact on performance. 
 
\subsubsection{Negative edges}
\label{sec:neg_edge_exp}

\begin{figure}
\centering
\includegraphics[width=0.9\columnwidth]{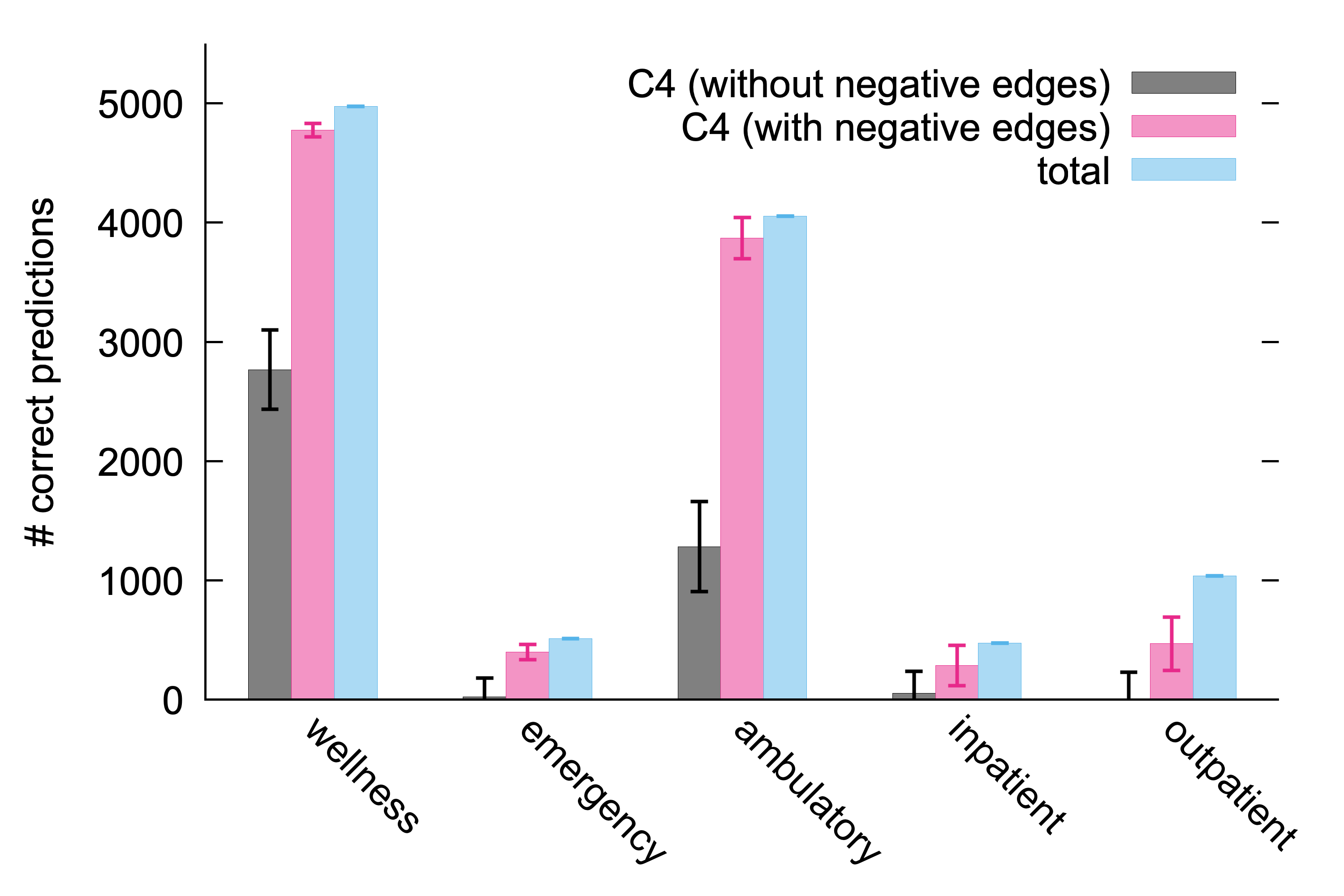}
\caption{Illustrating how the negative edges affect the prediction of care actions.  The results are the average of 50 random samples of the Synthea dataset. Each sample consists of 50/10 patients in the training/testing. The nodes' embeddings have size 5 and the GNN has 2 layers.}
\label{fig:negative_edges}
\end{figure}

This experiment studies the impact of removing negative edges in the KG. Recall that the negative edges in the KG are from the encounter to the care actions with no positive edge (see also Sec.\ \ref{sec:kg_construction}). We use the C4 GNN connectivity introduced in Sec.\ \ref{sec:gnn_connectivity} and obtain the results shown in
Figure \ref{fig:negative_edges}. The figure shows that not using negative edges considerably drops the link prediction performance. This behavior is because the link prediction can be thought of as a binary classification of an edge, and negative edges are a source of negative samples of such ``classification.'' Notably,  negative edges are not readily available in the Synthea generated data, and we had to use domain knowledge (i.e., understanding of the data) to add those. 

\textbf{Conclusions:} Negative edges are crucial for making good link predictions in this use case. The negative links were not available in Synthea directly, and we had to use domain knowledge (i.e., understanding of the graph data) to include them in the KG. 

%%%%%%

\section{Conclusions}
This paper studied the flow of embedding information within GNNs and its impact on performance, specifically in a clinical triage use case. We proposed a mathematical model that decouples the GNN connectivity from the connectivity of graph data and found that incorporating domain knowledge in the GNN connectivity is more effective than relying solely on graph data connectivity. Our results also show that negative edges play a crucial role in achieving good performance, while using many GNN layers can lead to performance degradation.

A future research direction is to evaluate how the approach performs in other datasets, and how to automatize the learning of the ``domain knowledge.'' Specifically, the identification of key GNN links for transporting embedding information, and the identification of negative edges in the KG that may not be explicitly present in the data. 
%These findings shed light on how to optimize GNN performance and provide valuable insights for future research in this field.
%%%%%%%
\bibliographystyle{IEEEtran}
\bibliography{references}

\begin{thebibliography}{10}
\providecommand{\url}[1]{#1}
\csname url@samestyle\endcsname
\providecommand{\newblock}{\relax}
\providecommand{\bibinfo}[2]{#2}
\providecommand{\BIBentrySTDinterwordspacing}{\spaceskip=0pt\relax}
\providecommand{\BIBentryALTinterwordstretchfactor}{4}
\providecommand{\BIBentryALTinterwordspacing}{\spaceskip=\fontdimen2\font plus
\BIBentryALTinterwordstretchfactor\fontdimen3\font minus
  \fontdimen4\font\relax}
\providecommand{\BIBforeignlanguage}[2]{{%
\expandafter\ifx\csname l@#1\endcsname\relax
\typeout{** WARNING: IEEEtran.bst: No hyphenation pattern has been}%
\typeout{** loaded for the language `#1'. Using the pattern for}%
\typeout{** the default language instead.}%
\else
\language=\csname l@#1\endcsname
\fi
#2}}
\providecommand{\BIBdecl}{\relax}
\BIBdecl

\bibitem{ZSM+20}
X.~Zeng, X.~Song, T.~Ma, X.~Pan, Y.~Zhou, Y.~Hou, Z.~Zhang, K.~Li, G.~Karypis,
  and F.~Cheng, ``Repurpose open data to discover therapeutics for covid-19
  using deep learning,'' \emph{Journal of proteome research}, vol.~19, no.~11,
  pp. 4624--4636, 2020.

\bibitem{BUG13}
A.~Bordes, N.~Usunier, A.~Garcia-Duran, J.~Weston, and O.~Yakhnenko,
  ``Translating embeddings for modeling multi-relational data,'' \emph{Advances
  in neural information processing systems}, vol.~26, 2013.

\bibitem{XHH+15}
H.~Xiao, M.~Huang, Y.~Hao, and X.~Zhu, ``Transg: A generative mixture model for
  knowledge graph embedding,'' \emph{arXiv preprint arXiv:1509.05488}, 2015.

\bibitem{ZCZ+20}
Z.~Zhang, J.~Cai, Y.~Zhang, and J.~Wang, ``Learning hierarchy-aware knowledge
  graph embeddings for link prediction,'' in \emph{Proceedings of the AAAI
  conference on artificial intelligence}, vol.~34, no.~03, 2020, pp.
  3065--3072.

\bibitem{Zhou2020}
J.~Zhou, G.~Cui, S.~Hu, Z.~Zhang, C.~Yang, Z.~Liu, L.~Wang, C.~Li, and M.~Sun,
  ``{Graph neural networks: A review of methods and applications},'' \emph{AI
  Open}, vol.~1, 2020.

\bibitem{SKB+18}
M.~Schlichtkrull, T.~N. Kipf, P.~Bloem, R.~Van Den~Berg, I.~Titov, and
  M.~Welling, ``Modeling relational data with graph convolutional networks,''
  in \emph{The Semantic Web: 15th International Conference, ESWC 2018,
  Heraklion, Crete, Greece, June 3--7, 2018, Proceedings 15}.\hskip 1em plus
  0.5em minus 0.4em\relax Springer, 2018, pp. 593--607.

\bibitem{Ahmedt21}
D.~Ahmedt-Aristizabal, M.~Ali~Armin, S.~Denman, C.~Fookes, and L.~Petersson,
  ``{Graph-Based Deep Learning for Medical Diagnosis and Analysis: Past,
  Present and Future},'' \emph{Sensors (Basel)}, vol.~14, 2021.

\bibitem{Siyi2023}
S.~Tang, A.~Tariq, J.~A. Dunnmon, U.~Sharma, P.~Elugunti, D.~L. Rubin, B.~N.
  Patel, and I.~Banerjee, ``Predicting 30-day all-cause hospital readmission
  using multimodal spatiotemporal graph neural networks,'' \emph{IEEE Journal
  of Biomedical and Health Informatics}, pp. 1--12, 2023.

\bibitem{Golmaei2021}
S.~N. Golmaei and X.~Luo, ``Deepnote-gnn: Predicting hospital readmission using
  clinical notes and patient network,'' in \emph{Proceedings of the 12th ACM
  Conference on Bioinformatics, Computational Biology, and Health Informatics},
  2021.

\bibitem{Baribiero2021}
H.~Lu and S.~Uddin, ``A weighted patient network-based framework for predicting
  chronic diseases using graph neural networks,'' \emph{Sci Rep}, vol.~11,
  2021.

\bibitem{Zhu2021}
W.~Zhu and N.~Razavian, ``Variationally regularized graph-based representation
  learning for electronic health records,'' in \emph{Proceedings of the
  Conference on Health, Inference, and Learning}, 2021.

\bibitem{Choi2020}
E.~Choi, Z.~Xu, Y.~Li, M.~W. Dusenberry, G.~Flores, E.~Xue, and A.~M. Da,
  ``Learning the graphical structure of electronic health records with graph
  convolutional transformer,'' \emph{AAAI}, 2020.

\bibitem{HYL17}
W.~Hamilton, Z.~Ying, and J.~Leskovec, ``Inductive representation learning on
  large graphs,'' \emph{Advances in neural information processing systems},
  vol.~30, 2017.

\bibitem{pyg}
\BIBentryALTinterwordspacing
 [Online]. Available: \url{https://www.pyg.org}
\BIBentrySTDinterwordspacing

\bibitem{PyG-introductory-example}
\BIBentryALTinterwordspacing
P.~Documentation. (2023) Introduction by example. [Online]. Available:
  \url{https://pytorch-geometric.readthedocs.io/en/latest/get\_started/introduction.html\#}
\BIBentrySTDinterwordspacing

\bibitem{LYK+20}
J.~Lee, W.~Yoon, S.~Kim, D.~Kim, S.~Kim, C.~H. So, and J.~Kang, ``Biobert: a
  pre-trained biomedical language representation model for biomedical text
  mining,'' \emph{Bioinformatics}, vol.~36, no.~4, pp. 1234--1240, 2020.

\bibitem{WKN+18}
J.~Walonoski, M.~Kramer, J.~Nichols, A.~Quina, C.~Moesel, D.~Hall, C.~Duffett,
  K.~Dube, T.~Gallagher, and S.~McLachlan, ``Synthea: An approach, method, and
  software mechanism for generating synthetic patients and the synthetic
  electronic health care record,'' \emph{Journal of the American Medical
  Informatics Association}, vol.~25, no.~3, pp. 230--238, 2018.

\bibitem{KB14}
D.~P. Kingma and J.~Ba, ``Adam: A method for stochastic optimization,''
  \emph{arXiv preprint arXiv:1412.6980}, 2014.

\bibitem{CLL+20}
D.~Chen, Y.~Lin, W.~Li, P.~Li, J.~Zhou, and X.~Sun, ``Measuring and relieving
  the over-smoothing problem for graph neural networks from the topological
  view,'' in \emph{Proceedings of the AAAI conference on artificial
  intelligence}, vol.~34, no.~04, 2020, pp. 3438--3445.

\bibitem{CW20}
C.~Cai and Y.~Wang, ``A note on over-smoothing for graph neural networks,''
  \emph{arXiv preprint arXiv:2006.13318}, 2020.

\end{thebibliography}

\end{document}